\pdfoutput=1

\documentclass[11pt]{article}

\usepackage[final]{acl}

\usepackage{times}
\usepackage{latexsym}
\usepackage{algorithm} 
\usepackage{algorithmic}
\usepackage{graphicx}
\usepackage{makecell,multirow,diagbox}
\usepackage{amsmath}
\usepackage{colortbl}

\usepackage[T1]{fontenc}

\usepackage[utf8]{inputenc}

\usepackage{microtype}

\usepackage{inconsolata}

\title{Exploring Mathematical Extrapolation of \\Large Language Models with Synthetic Data}

\author{Haolong Li\thanks{Work done during internship at ByteDance.}\\
  Tongji Universiy\\
  \texttt{furlongli322@gmail.com} 
  \And
  Yu Ma \\ 
  Seed Foundation, ByteDance \\
  \texttt{mayu.1231@bytedance.com} 
  \And
  Yinqi Zhang$^*$ \\
  East China Normal University\\
  \texttt{zhang.inch@gmail.com}\\
  \AND
  Chen Ye$^\dagger$ \\
  ESSC Lab, Tongji Universiy\\
  \texttt{yechen@tongji.edu.cn}\\
  \And
  Jie Chen\thanks{Corresponding Author}\thanks{Project Leader} \\
  Seed Foundation, ByteDance\\
  \texttt{chenjiexjtu@gmail.com}
  }

\begin{document}
\maketitle
\begin{abstract}
Large Language Models (LLMs) have shown excellent performance in language understanding, text generation, code synthesis, and many other tasks, while they still struggle in complex multi-step reasoning problems, such as mathematical reasoning. In this paper, through a newly proposed arithmetical puzzle problem, we show that the model can perform well on multi-step reasoning tasks via fine-tuning on high-quality synthetic data. Experimental results with the open-llama-3B model on three different test datasets show that not only the model can reach a zero-shot pass@1 at 0.44 on the in-domain dataset, it also demonstrates certain generalization capabilities on the out-of-domain datasets. Specifically, this paper has designed two out-of-domain datasets in the form of extending the numerical range and the composing components of the arithmetical puzzle problem separately. The fine-tuned models have shown encouraging performance on these two far more difficult tasks with the zero-shot pass@1 at 0.33 and 0.35, respectively.
\end{abstract}
\section{Introduction}

Large Language Models (LLMs), as zero-shot and multi-task learners, have shown extraordinary capabilities across a variety of natural language tasks \citep{vaswani2017attention,schulman2017proximal,radford2019language,ziegler2019fine, brown2020language,kojima2022large,park2023generative, chowdhery2023palm, rafailov2024direct}. However, even the most advanced LLMs face challenges when it comes to tackling complex multi-step reasoning problems, such as mathematical and scientific reasoning \citep{koncel2016mawps, cobbe2021training, hendrycks2021measuring, wei2022chain, chen2022program, gao2023pal, trinh2024solving}. This comes from three main reasons: firstly, mathematical reasoning often requires quantitative multiple steps of deduction, since a single logical error is enough to derail a much larger solution \citep{lightman2023let}. Secondly, the lack of high-quality data limits LLMs' ability to generalize and excel in mathematical reasoning tasks. Lastly, LLMs encounter difficulty in extrapolation, as they struggle to apply reasoning skills when solving unseen mathematical problems.

Many prior research has explored along these challenges. GPT-4 \citep{achiam2023gpt}, LLaMA \citep{touvron2023llama, touvron2023llama1}, Gemini \citep{team2023gemini}, Minerva \citep{lewkowycz2022solving}, Llemma \citep{azerbayev2023llemma}, Mistral \citep{jiang2023mistral}, WizardMath \citep{luo2023wizardmath}, MAMMOTH \citep{yue2023mammoth}, ToRA \citep{gou2023tora} and Deepseek \citep{bi2024deepseek,guo2024deepseek,lu2024deepseek} have emerged as dominant models in popular mathematical reasoning benchmarks such as GSM8K \citep{cobbe2021training}, MATH \citep{hendrycks2021measuring}, CMATH \citep{wei2023cmath} and AGIEval \citep{zhong2023agieval}. Moreover, process supervision and verifiers \citep{cobbe2021training, li2023making, uesato2022solving, lightman2023let, yu2023outcome} at the step level have also obtained widespread attention. However, mathematical extrapolation, particularly in terms of abstract forms, is often overlooked.

In this paper, we address the aforementioned challenges by introducing a novel and challenging arithmetical puzzle problem and making an initial attempt to solve them. Specifically, we propose a puzzle that needs multi-step calculations to generate a correct solution. Meanwhile, a data synthesis pipeline is developed to automatically generate a vast amount of high-quality data for supervised fine-tuning (SFT). And a series of LLMs based on open-llama-3B \citep{touvron2023llama} are fine-tuned on this synthetic dataset. Furthermore, to demonstrate the reasoning abilities in extrapolation, we have designed two out-of-domain benchmarks in the form of extending the numerical range and the composing components of the arithmetical puzzle problem. For the purpose of fair evaluation, we have restricted our models to greedy sampling in a zero-shot setting and provided a corresponding verifier. Our data scaling experiments demonstrate that as the amount of synthetic data grows, in-domain zero-shot pass@1 increases from 0.22 to 0.44, while the out-of-domain zero-shot pass@1 increases from 0.14/0.17 to 0.33/0.35.

Our major contributions can be concluded as: (1) We propose a novel arithmetical puzzle problem with corresponding data synthesis pipeline and out-of-domain benchmarks, to verify the multi-step reasoning and extrapolation capabilities of LLMs fine-tuned on synthetic data. (2) Experiments indicate that increasing the amount of high-quality synthetic data leads to performance enhancements across in-domain and out-of-domain datasets. (3) A comprehensive case study has been performed.

 \begin{table}[t]
    \centering
    \resizebox{\linewidth}{!}{
    \begin{tabular}{|l|}
    \hline
    \textbf{Example of the Synthetic Data}\\
    \hline 
    \makecell[l]{
    \textbf{---prompt---}\\
    34, 18, 31, 41, 19, 55: -110\\
    \\
    \textbf{---response---}\\
    31-34=-3, 19+41=60, 60/-3=-20, -20/18=-2, \\
    -2*55=-110
    }\\
    \hline
    \end{tabular}
    }

\caption{Example of our synthetic data.} 
    \vspace{-1.5em}
\label{tab:demo_prompt_resp}
\end{table}

\section{Problem Definition}

\subsection{Arithmetical Puzzle Problem} \label{sec:prob_desc}

Arithmetical puzzle problem denotes a mathematical puzzle involving arithmetic operations and requires logical reasoning and numerical manipulation to derive a solution. The 24 Puzzle and Arithmetic Grid Puzzle are well-known examples of arithmetical puzzle problems.

In this paper, we propose a challenging arithmetical puzzle. Its objective is intricate yet precise: to deftly manipulate a set of given integers through a calculated sequence of arithmetic operations, to achieve a predetermined target integer. The problem strictly limits each integer to be used by one time exactly. For example, for the integers $3, 6, 7, 51, 58$ and the target integer $4$, one possible solution is: $58-51=7$, $6-7=-1$, $3\times-1=-3$, $-3+7=4$, as shown in Figure~\ref{fig:main} in Appendix~\ref{sec:pv}.

\subsection{Data Synthesizing}
Given the arithmetical puzzle described above in Section~\ref{sec:prob_desc}, we create a data synthesizing pipeline to efficiently generate the proposed dataset. 

Denote the set of candidate integers as $X=\{X_1, X_2, \ldots, X_N\}$ and the target number as $T$, where $N$ is the total number of candidate integers in a puzzle sample. Each candidate integer $X_i$ is independently sampled from a uniform distribution $X_i \sim \text{U}(1, V)$, where $V$ is the upper bound of the sampled integers. To avoid data overlapping, we have strictly ensured that for each puzzle, the candidate integers are a set of distinct numbers. The arithmetic operators involved in this problem are $ops = \{+, -, \times, \div \}$ and all operations are limited to integer operations. For example, when solving the puzzle with a division operator, the operation should be considered in integer division like $14/3=4$. The detailed steps of synthesizing data for this puzzle is described in Algorithm~\ref{alg:algorithm1}.

Besides, to construct the SFT dataset, the prompt is deliberately designed to excludes any natural language cues and instead focuses on purely symbolic language. See Table~\ref{tab:demo_prompt_resp} for an example of the constructed prompt and response.

\begin{algorithm}[t]
\caption{Data Synthesis Algorithm}
\label{alg:algorithm1}
\begin{algorithmic}[1]
\STATE $S_{dataset}$ starts with an empty set
\WHILE{$size_{S_{dataset}} \leq size_{limit}$}
\STATE Sample $\{X_i \mid1 \leq i \leq N, X_i \sim \text{U}(1, V)\}$
\STATE $L$ starts with an empty list
\STATE $S \gets \{X_i\}$
\FOR{$i = 1$ to $N-1$}
\STATE Randomly select $a_i, b_i \in S$
\STATE Randomly select $ops_i \in \{+,-,\times,\div\}$
\STATE $c_i \gets a_i\quad ops_i\quad b_i$
\STATE $S \gets S-\{a_i\}-\{b_i\}$
\STATE $S \gets S \cup \{c_i\}$
\STATE $L \gets L + \{a_i. ops_i. b_i, c_i\}$
\ENDFOR
\STATE $T \gets c_{N-1}$
\IF{$\{L,T\} \notin S_{dataset}$}
\STATE $S_{dataset} \gets S_{dataset} \cup \{L,T\}$
\ENDIF
\ENDWHILE
\end{algorithmic}
\end{algorithm}

\subsection{Dataset} \label{sec:dataset}
\begin{figure*}[htb]
 \begin{minipage}{0.48\linewidth}
       \centering
   \includegraphics[width=0.94\textwidth]{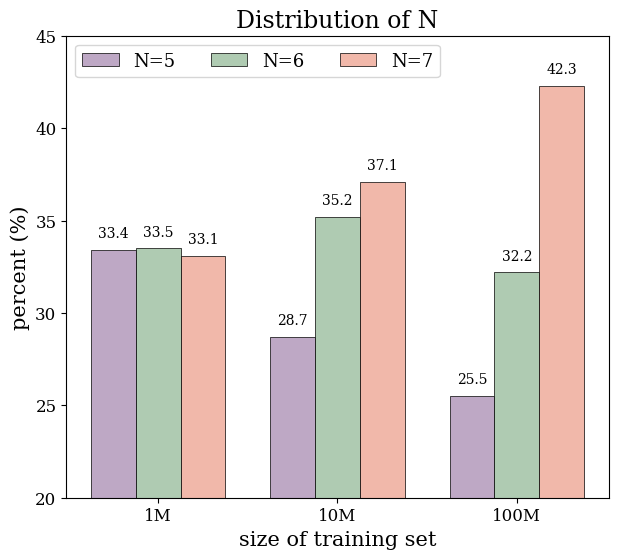}
   \label{subfig:dn}
   
 \end{minipage}
 \hfill 
 \begin{minipage}{0.48\linewidth}
   \centering
   \includegraphics[width=0.94\textwidth]{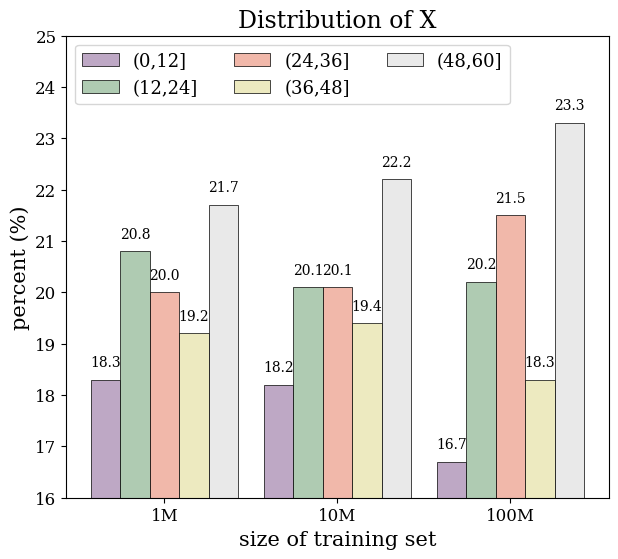}
   \label{subfig:dx}
   
 \end{minipage}
 \caption{Distributions of $N$ and $X$ for different training set sizes (1M / 10M / 100M samples). $N$ denotes the total number of candidate integers of our puzzle, $X = (X_1, X_2, \ldots, X_N)$ denotes the candidate integers.}
 \label{fig:data_dist}
 \end{figure*}

\begin{figure*}[htb]
 \begin{minipage}{0.48\linewidth}
       \centering
   \includegraphics[width=\textwidth]{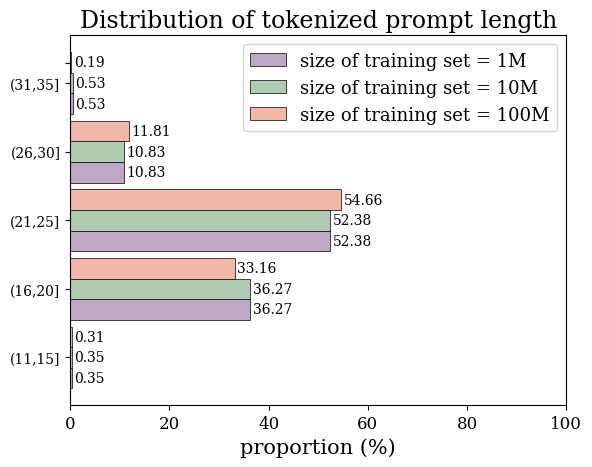}
   \label{subfig:DPL}
   
 \end{minipage}
 \hfill 
 \begin{minipage}{0.48\linewidth}
   \centering
   \includegraphics[width=\textwidth]{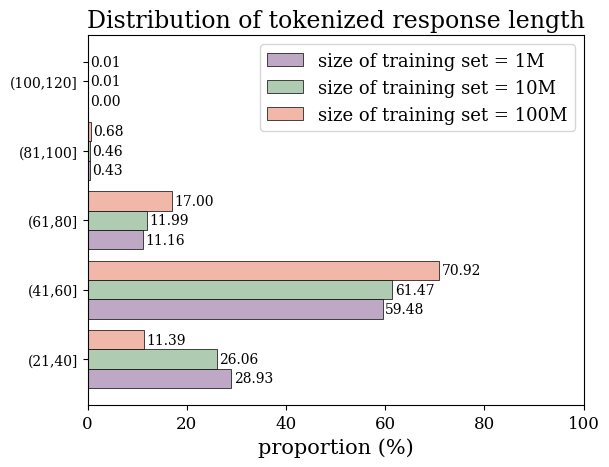}
   \label{subfig:DTL}
   
 \end{minipage}
    \vspace{-1.em}
 \caption{Distributions of the tokenized prompt and response lengths for different training set sizes (1M / 10M / 100M samples).}
    \vspace{-1.5em}
 \label{fig:prompt-resp-len}
 \end{figure*}

We split the dataset into training and in-distribution and out-of-distribution test dataset by controlling the total number of candidate integers $N$ and the upper bound of the sampled integers $V$. We set $ V = 60$ for the training dataset, and sampled the candidate integers with $N = 5, 6, 7$. Three training datasets with different sizes scaling from 1 million to 10 millions and 100 millions are generated. And another 7500 samples (2500 samples for each $N$) under the same setting are generated as the in-distribution test dataset. Figure.~\ref{fig:data_dist} shows the distribution of $N$ and $X$ in these three training datasets. And the corresponding distribution of the tokenized prompt and response length is shown in Figure.~\ref{fig:prompt-resp-len}.

\begin{figure*}[htb]
 \begin{minipage}{0.48\linewidth}
       \centering
   \includegraphics[width=0.94\textwidth]{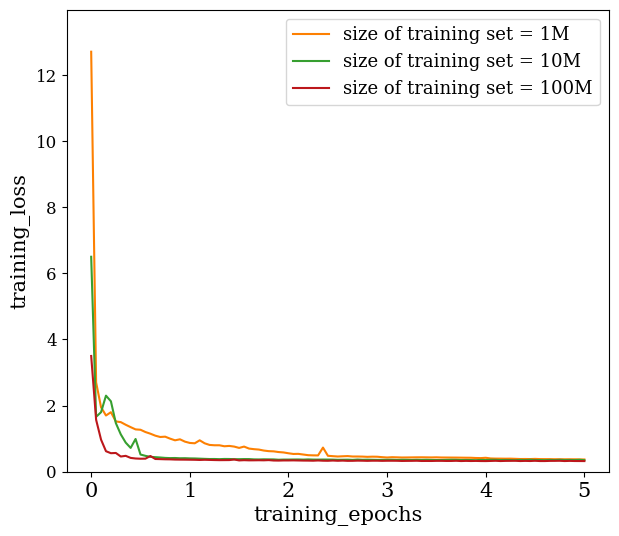}
   \label{subfig:loss}
   
 \end{minipage}
 \hfill 
 \begin{minipage}{0.48\linewidth}
   \centering
   \includegraphics[width=0.94\textwidth]{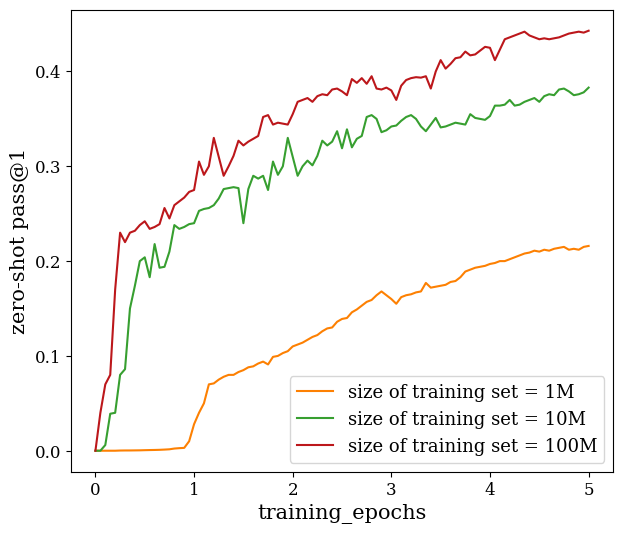}
   \label{subfig:zeroshot}
   
 \end{minipage}
 \caption{The training loss and zero-shot pass@1 on ID dataset for different training set sizes (1M / 10M / 100M samples).}
 \label{fig:training_curve}
\end{figure*}

To further evaluate the model's performance on extrapolation, we have also designed two benchmarks of out-of-distribution dataset:

\begin{algorithm}[tb]
\caption{Verifier Algorithm}
\label{alg:verifier}
\begin{algorithmic}[1]
\STATE $\{X_i \mid1 \leq i \leq N\} \gets X_{prompt}$
\STATE $T \gets T_{prompt}$
\STATE $Eqs \gets Solution_{generated}$
\STATE $S \gets \{X_i\}$
\STATE $Flag_{verifier} \gets False$
\FOR{$eq_i \in Eqs$}
\IF{$eq_i$ is a legel equation}
\STATE $a_i, ops_i, b_i, c_i \gets ParseEq(eq_i)$
\IF{$a_i, b_i \in S$}
\STATE $S \gets S-\{a_i\}-\{b_i\}$
\STATE $S \gets S \cup \{c_i\}$
\ELSE
\STATE $break$
\ENDIF
\ELSE
\STATE $break$
\ENDIF
\ENDFOR
\IF{$c_{N-1} = T$}
\STATE $Flag_{verifier} \gets True$
\ENDIF
\end{algorithmic}
\end{algorithm}
\textbf{Numerical OOD test datasets}. The upper bound of the sampled integers $V$ is raised to 100 and 1000 separately to test the model's generalization ability with unseen larger numbers. Specifically, 6000 samples are generated for each value of $V$ with 2000 samples for each $N$. An additional filtering pipeline is applied to ensure that for each sample, there exists at least one integer $X_i$ that satisfies $60 < X_i < 100$ for the dataset with $V = 100$ and $100 < X_i < 1000$ for that with $V = 1000$.

\textbf{Form OOD test dataset}. In mathematics, abstract forms often extend, such as expanding from a two-variable linear equation to one with three variables. For the proposed arithmetic puzzle, the extrapolation of abstract forms can be achieved by changing the number of candidate integers $N$. Clearly, when $N$ increases, the exploration space leading to a feasible solution would expand exponentially, which results in an increased demand for precise reasoning steps. From another perspective, when the total number of the candidate integers changes, it actually requires the model's ability to absorb and adapt to the puzzle's abstract forms. Therefore, to test the model's generalization capability from this point of view, we create another benchmark for OOD test dataset with 5000 samples generated with setting $N$ to 8. To control variables, all the candidate integers in this dataset are sampled with the same upper bound $V = 60$ as the training dataset.

\section{Model}
\subsection{Framework}

We adopt the llama architecture \citep{touvron2023llama} and employ low-rank adaptation (LoRA) tuning \citep{hu2021lora} based on the implementation of TRL full stack library \citep{vonwerra2022trl}. 
LoRA achieves a remarkable reduction of 89\% in our trainable parameters, from 3B to 0.3B.


\subsection{Implementation Details}
We train our model by fine-tuning open-llama-3B. We systematically apply left-padding to the query text and right-padding to the answer text to control the overall context length. All experiments are conducted with $8\times$ NVIDIA A100-SXM4-80GB GPUs. The specific hyperparameter settings are listed in Table~\ref{tab:1} in Appendix~\ref{sec:settings}.


\section{Experiments}

\subsection{Evaluation}

For the fine-tuned model, we use the greedy decoding strategy in a zero-shot setting to generate responses. To measure the model's performance on the proposed puzzle, a corresponding verifier is designed to automatically evaluate the correctness of the responses. Specifically, a solution is deemed correct if it satisfies the following rules:
\begin{itemize}
    \item No extra or illegal characters.
    \vspace{-0.8em}
    \item There are only $N-1$ equations and all the corresponding calculations are correct.
    \vspace{-0.8em}
    \item $F$($X_1$,\quad \ldots,\quad $X_N$ $\mid$ $ops$) = $T$.
    \vspace{-0.8em}
    \item All $\{X_i\mid i\in\{1, 2, \ldots, N\}\}$ and the intermediate calculation results are only used once.
\end{itemize}
The detailed steps of evaluating the solution for this puzzle is described in Algorithm~\ref{alg:verifier}.

\subsection{Results}

As mentioned in Section~\ref{sec:dataset}, we have generated three training datasets with different sizes to explore the data scaling effects on the fine-tuned model. The pass@1 rate on different in-distribution and out-of-distribution test datasets are shown in Table~\ref{tab:pass1}. When the model is fine-tuned with 100M samples, it achieves the highest score with a zero-shot pass@1 of 0.44 in the in-distribution test dataset, and 0.33 and 0.35 in the two OOD datasets, respectively.

Furthermore, we have shown the training curves of the model fine-tuned on these three datasets in Figure~\ref{fig:training_curve}. From Figure~\ref{fig:training_curve}, a faster decaying rate is clearly observed in the training loss when increasing the training data size, which is consistent with the rapid increase of the pass@1 rate evaluated on the in-distribution dataset. The same enhancement of the performance also occurs in the two OOD test datasets as shown in Table \ref{tab:pass1}.

Additionally, we have also conducted tests of this puzzle on the base model (open-llama-3B) and several other open-source and closed-source models with both few-shot and CoT prompting. The results and some of the generated cases are shown in Appendix~\ref{appendix:base-eval}, demonstrating the necessity of fine-tuning with regard to solving such puzzle problems.

 \begin{table*}[htb]
    \centering
    \resizebox{\linewidth}{!}{
    \begin{tabular}{|c|c|c|c|c|c|c|}
    \hline
    \multicolumn{1}{|c|}{\textbf{Dataset}} &\textbf{Range}&\textbf{Number of Integers}&\textbf{Fine-tuned on 1M}   & \textbf{Fine-tuned on 10M}  & \textbf{Fine-tuned on 100M}\\
    \hline
    \multirow{3}{*}{ID}  &\multirow{3}{*}{[1,60]}&5& 0.224 & 0.428 & \textbf{0.471} \\
    &&6& 0.208 & 0.363 & \textbf{0.432} \\
    &&7& 0.205 & 0.360 & \textbf{0.425} \\
    \arrayrulecolor{gray!50}
    \hline
    \arrayrulecolor{black}
    Total ID  &[1,60]&5,6,7& 0.216 & 0.383 & \textbf{0.443} \\
    \hline
    \multirow{3}{*}{Numerical OOD}&\multirow{3}{*}{[1,100]}&5& 0.163 & 0.239 & \textbf{0.364} \\
    &&6& 0.137 & 0.199 & \textbf{0.331} \\
    &&7& 0.126 & 0.186 & \textbf{0.315} \\
    \arrayrulecolor{gray!50}
    \hline
    \arrayrulecolor{black}
    Total Numerical OOD&[1,100]&5,6,7 & 0.141 & 0.205 & \textbf{0.326} \\
    \hline
    \multirow{3}{*}{Numerical OOD}&\multirow{3}{*}{[1,1000]}&5&0.131	&0.181	& \textbf{0.229} \\
    &&6&0.030	&0.051	& \textbf{0.063} \\
    &&7&0.111	&0.163	& \textbf{0.220} \\
    \arrayrulecolor{gray!50}
    \hline
    \arrayrulecolor{black}
    Total Numerical OOD&[1,1000]&5,6,7&0.091	&0.132	& \textbf{0.170} \\
    \hline
    Form OOD&[1,60]&8& 0.169 & 0.231 & \textbf{0.352} \\
 
    \hline
    \end{tabular}
    }

\caption{Zero-shot pass@1 of the model fine-tuned with different training set sizes (1M / 10M / 100M samples) on ID, numerical OOD, and form OOD test datasets. The best results are \textbf{highlighted}.} 
    \vspace{-1.em}
\label{tab:pass1}
\end{table*}

\subsection{Case Studies}

We further demonstrate the different solutions provided by models trained with 1M / 10M / 100M training data on the form OOD test dataset for several challenging queries. As shown in Figure~\ref{fig:cases} in Appendix~\ref{sec:cases}, the model trained on 1M samples is still limited to a fixed number of reasoning steps, whereas the models trained on 10M / 100M samples exhibit a higher-level understanding of the problem and perform an adequate number of reasoning steps. However, compared to the model trained on 100M samples, the model trained on 10M samples may still encounter computational or logical errors in the final step of reasoning. 


\section{Conclusion}
Large language models (LLMs) are intrinsically zero-shot and multi-task learners. However, mathematical reasoning still poses challenges for LLMs, we propose that the reasons can be mainly categorized into three folds: (1) Requirement of multi-step derivation; (2) Lack of high quality data for fine-tuning; (3) Difficulty in extrapolation. In this paper, we design an arithmetical puzzle and make an early attempt to solve these challenges. We develop a 24-point puzzle-like problem which asks for multi-step calculations to arrive at the correct answer. A corresponding data synthesis pipeline is proposed to generate an arbitrary amount of high-quality data, on which a series of LLMs are fined-tuned. In order to verify the extrapolation capability of our models, we have designed two out-of-domain benchmarks and show that our model achieves competitive performance. Furthermore, a data scaling experiment is conducted and it is concluded that by increasing the amount of training data, both the training loss and in/out-of-domain performance of the fine-tuned model improve accordingly.

\section*{Acknowledgements}
We appreciate Peng Sun for providing the initial SFT dataset, and Xintian Han for suggestions about the reward calculation and ablation study. We would also like to thank Liang Xiang and Xun Zhou for the helpful discussions across the project.

\newpage
\section{Limitations}

In this study, we have explored the mathematical extrapolation of Large Language Models (LLMs) and discovered that, with high-quality synthetic data, LLMs demonstrates certain generalization capabilities in mathematical extrapolation. However, LLMs have not yet fully mastered this capability, and it remains uncertain if this ability can be extended to other complex mathematical tasks. In the future, our research will focus on investigating and enhancing this capability, aiming to empower LLMs to explore unsolved mathematical problems through leveraging our existing knowledge.

\section{Ethics Statement}
In this research, we adhere to strict ethical guidelines and principles. The study has been designed and implemented with respect for the rights, privacy, and well-being of all individuals involved. All of our data is synthesized using our proposed data synthesis algorithm, ensuring compliance with relevant regulations and standards. Our findings and conclusions are reported accurately and objectively, avoiding any misrepresentation or manipulation of data. The entire process and outcomes are free from intellectual property and ethical legal disputes.

\bibliography{custom}

\newpage
\appendix
\onecolumn

\section{Appendix}

\subsection{Hyperparameter Settings}
\label{sec:settings}

In the SFT stage, we follow common fine-tuning hyperparameter settings for our model. We set learning rate to $1e-4$ and adopt the cosine learning rate scheduler. We use low-rank adaptation (LoRA) tuning with a rank of $5$, $\alpha$ of $32$, and dropout of $0.05$. And we employ Adamw optimizer with $\beta1=0.9$, $\beta2=0.95$ and $\epsilon=1e-9$. Eight NVIDIA A100-SXM4-80GB GPUs are used to train the model with a batch size of $50$ and the maximum epoch set to $5$.  Detailed settings are listed in Table~\ref{tab:1}.

 \begin{table*}[htb]
    \centering
    \begin{tabular}{|c|c|c|c|}
    \hline
    \textbf{Hyperparameter}   & \textbf{Value} &\textbf{Hyperparameter}    & \textbf{Value} \\
    \hline
    Learning Rate & $1e-4$ &  Epochs & $5$ \\
    Learning Rate Scheduler & Cosine& Optimizer & Adamw\\
    Warmup Step & $0$ & Optimizer $\beta1$ & $0.9$\\
    GPU Nums & $8$ & Optimizer $\beta2$ & $0.95$\\
    Batch Size Per GPU & $50$ & Optimizer $\epsilon$ & $1e-9$\\
    Max Query Length & $36$ & Precision & AMP\\
    Max Answer Length & $130$ & LoRA Rank & $8$\\
    Max Generated Length & $167$ & LoRA $\alpha$ & $32$\\
    Precision & bfloat16  & LoRA Dropout & $0.05$\\
    Vocabulary Size & $32002$ & Seed & $1234$\\
    \hline
    \end{tabular}
\caption{Hyperparameter Settings.}
\vspace{-0.5em}

\label{tab:1}
\end{table*}
\subsection{Evaluation of the Base Model} \label{appendix:base-eval}
We evaluate the base model (open-llama-3B) on the proposed arithmetical puzzle problem. As shown in Table~\ref{tab:new1} and Table~\ref{tab:cot}, with either the few-shot prompting (2-Shot, 8-Shot) or Chain-of-Thought (CoT), the base model performs poorly on the puzzle. We propose this is due to the symbolic form of our prompt, the model needs to understand the underlying pattern in order to solve the arithmetical puzzle. Without fine-tuning on the synthetic data, the model may struggle to comprehend such type of prompt.

 \begin{table*}[htb]
    \centering
    \begin{tabular}{|l|c|c|c|c|}
    \hline
     \multicolumn{1}{|c|}{\textbf{Approach}}  &\textbf{ID}   & \textbf{OOD-N}  & \textbf{OOD-F} & \textbf{OOD-1000}\\
    \hline
    2-Shot Prompting& 0& 0& 0& 0 \\
    8-Shot Prompting& 0& 0& 0& 0 \\
    Chain-of-Thought & 0& 0& 0& 0 \\ 
    \hline
    \end{tabular}
\caption{Evaluation of the base model with few-shot and Chain-of-Thought prompting. As expected, the base model performs poorly across all the prompting techniques.} 
\label{tab:new1}
\end{table*}

 \begin{table*}[htb]
    \begin{tabular}{|l|l|}
    \hline
     \multicolumn{1}{|c|}{\textbf{Chain-of-Thought Prompting}}&\multicolumn{1}{c|}{\textbf{Generated Response of the Base Model}}\\
    \hline
    Let's play an arithmetic puzzle game. & The first thing to do is to find the number of ways to get 8.\\

    First, we are given a set of numbers: &
    There are 4 ways to get by adding 8 to 37, 11, 26, 15.\\
    
    8, 37, 11, 26, 15. The objective is to &There are 4 ways to get 8 by subtracting 8 from 37, 11, 26, 15.\\
    
    use operations like addition, &There are 4 ways to get 8 by multiplying 8 from 37, 11, 26, 15.\\
    
    subtraction, multiplication, and &There are 4 ways to get 8 by dividing 8 from 37, 11, 26, 15.\\
    
    division to obtain the number 8, &There are 4 ways to get 8 by adding 8 from 37, 11, 26, 15.\\
    using each number exactly once. &There are 4 ways to get 8 by subtracting 8 from 37, 11, 26, 15.\\
    Please provide the reasoning &There are 4 ways to get 8 by multiplying 8 from 37, 11, 26, 15.\\
    process step by step.&There are 4 ways to get 8 by dividing 8 from 37, 11, 26, 15.\\
    &\ldots\\
    
    
    
    \hline
    \end{tabular}

\caption{An example of Chain-of-Thought prompting and the generated response of the base model.} 
\label{tab:cot}
\end{table*}

We further test several open-source (Llama-2-7B \citep{touvron2023llama}, Deepseek-Coder-33B \citep{guo2024deepseek}) and closed-source models (GPT4 \citep{achiam2023gpt}) with few-shot prompting. As shown in Table~\ref{tab:new2}, these models also perform poorly on our benchmarks. In Table~\ref{tab:llms}, we provide an example of the CoT prompting and the generated responses from these models.


 \begin{table*}[htb]
    \centering
    \begin{tabular}{|c|c|c|c|c|}
    \hline
     \textbf{Model}&\textbf{ID}   & \textbf{OOD-N}  & \textbf{OOD-F} & \textbf{OOD-1000}\\
    \hline
    GPT4 & 0.002& 0.001& 0& 0.001 \\ 
    Llama-2-7B & 0& 0& 0& 0 \\
    Deepseek-Coder-33B & 0.001& 0& 0.001& 0 \\ 
    \hline
    \end{tabular}

\caption{Evaluation results of Llama-2-7B, Deepseek-Coder-33B, and GPT4 on our proposed benchmarks.} 
\label{tab:new2}
\end{table*}

 \begin{table*}[!htb]
    \resizebox{\columnwidth}{!}{
    \begin{tabular}{|c|c|c|}
    \hline
    \textbf{Few-Shot Prompting}& \textbf{Model} & \textbf{Response}\\
    \hline   
    \multirow{3}{*}{\makecell[l]{A puzzle problem. The first \\two lines are examples. Please \\solve the third line: \\36, 32, 57, 55, 11:30S11/36=0, \\0+32=32, 55+32=87, 87-57=30\\17, 6, 20, 48, 30:-40S17+48=65, \\30/6=5, 20+5=25, 25-65=-40 \\8, 37, 11, 26, 15:8S}} & GPT4 &
    \makecell[l]{Sure, I'll solve the third line of the puzzle using the pattern\\established in the first two lines:\\Firstly, divide the fifth number by the first number, which will \\yield an integer part:\\15 / 8 = 1 (ignore the decimal part)\\Next, add the result to the second number:\\1 + 37 = 38\\After that, add this result to the fourth number:\\38 + 26 = 64\\Final step, subtract the third number from this result:\\64 - 11 = 53\\However, the expected output provided is 8, but following the \\pattern above, the answer seems to be 53. It's possible that there \\are some other rules or factors at play, or therecould be an error \\in the input or expected output.}
    \\
    \cline{2-3}
    &Llama-2-7B& \makecell[l]{Sure, I'd be happy to help you solve the third line of the puzzle!\\The equation is:8,37,11,26,15:8S\\To solve this equation, we can start by simplifying the fraction:\\8/8 = 1\\So, the solution to the equation is:\\11,26,15 = 52\\I hope this helps! Let me know if you have any other questions.}\\
    \cline{2-3}
    &\makecell{DeepSeek-\\Coder-33B}& \makecell[l]{The third line of the puzzle problem is:\\17, 6, 20, 48, 30:-40S17+48=65, 30/6=5, 20+5=25, 25-65=-40\\The operations and numbers are:\\1. 17 + 48 = 65\\2. 30 / 6 = 5\\3. 20 + 5 = 25\\4. 25 - 65 = -40\\Let's solve these operations:\\1. 17 + 48 = 65\\2. 30 / 6 = 5\\3. 20 + 5 = 25\\4. 25 - 65 = -40 \\So, the solution to the third line is:\\17 + 48 = 65\\30 / 6 = 5\\20 + 5 = 25\\25 - 65 = -40\\This line of the puzzle problem is solved.} \\
    \hline
    \end{tabular}
    }

\caption{An example of few-shot prompting and the generated responses of GPT4, Llama-2-7B, and Deepseek-Coder-33B. We provide the models with two examples before the puzzle. As shown, all of the models fail to solve the given problem. GPT4 seems to understand the requirement of the puzzle, while the other two fail.} 
\label{tab:llms}
\end{table*}

As shown in Table~\ref{tab:llms}, Llama-2-7B fails to understand the requirement of the puzzle and just outputs two meaningless equations. Deepseek-Coder-33B treats the second example in few-shot prompting as the puzzle, and repeats the same calculations three times. It seems that GPT4 has well understood the prompt and used all the candidate integers only once, the calculations within the generated response are all right, while the solution is wrong. Actually, such kind of problem is very challenging, as the model needs to infer the requirement of the puzzle from the provided examples and then figure out the correct solution.

\subsection{Case Study}
\label{sec:cases}

\begin{figure*}[htb]
    \centering
    \includegraphics[width=0.97\columnwidth]{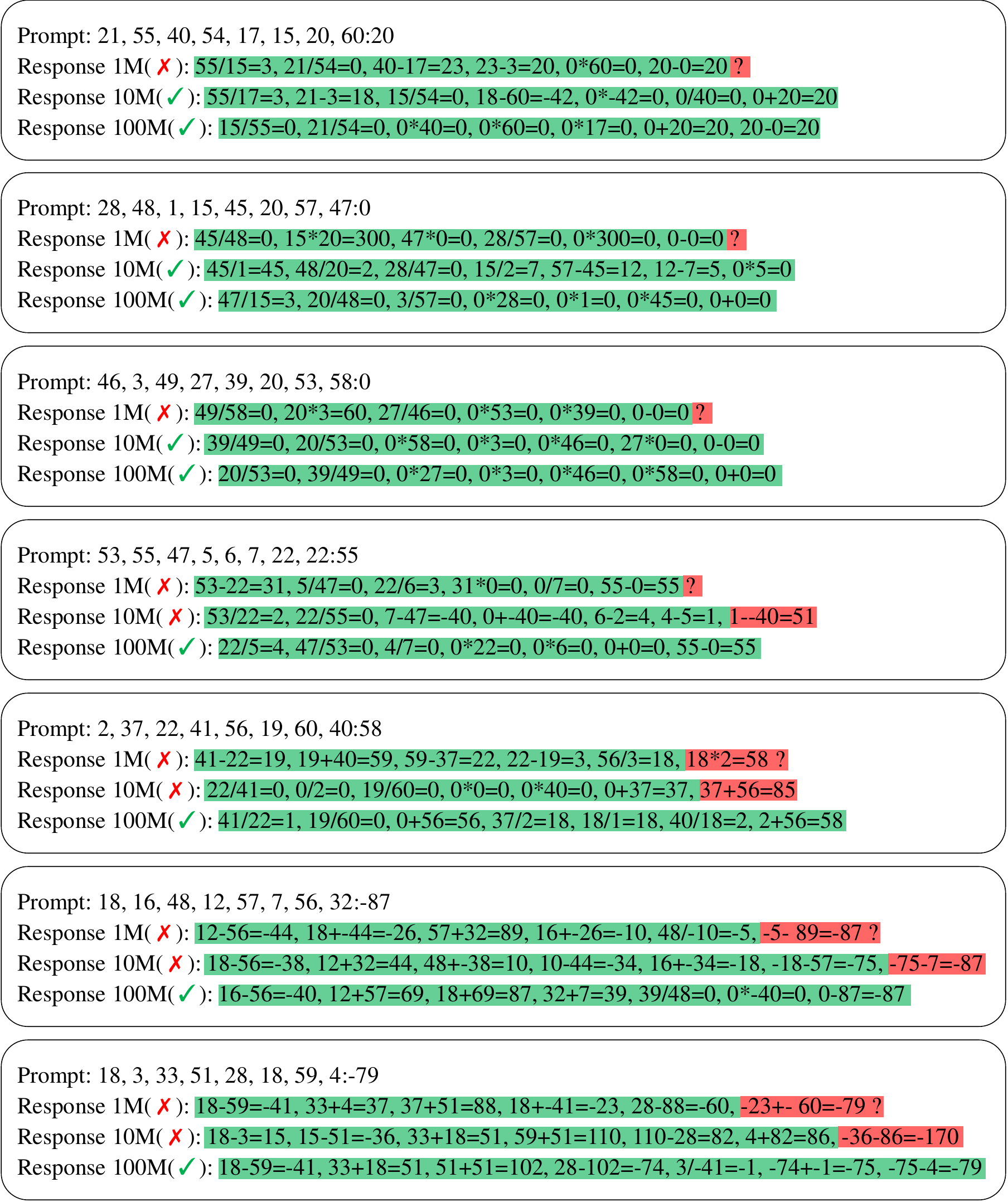}
    \caption{Cases from the form OOD test dataset. The correct steps are highlighted in green, while the incorrect steps in red. Generally speaking, performance of model fine-tuned with 1M training data is the worst.}
    \vspace{-1.2em}
    \label{fig:cases}
\end{figure*}
\vspace{1.5em}


    


\newpage
\subsection{Visualization of the Proposed Puzzle}
\label{sec:pv}

\begin{figure}[htb]
    \centering
    \includegraphics[width=0.6\columnwidth]{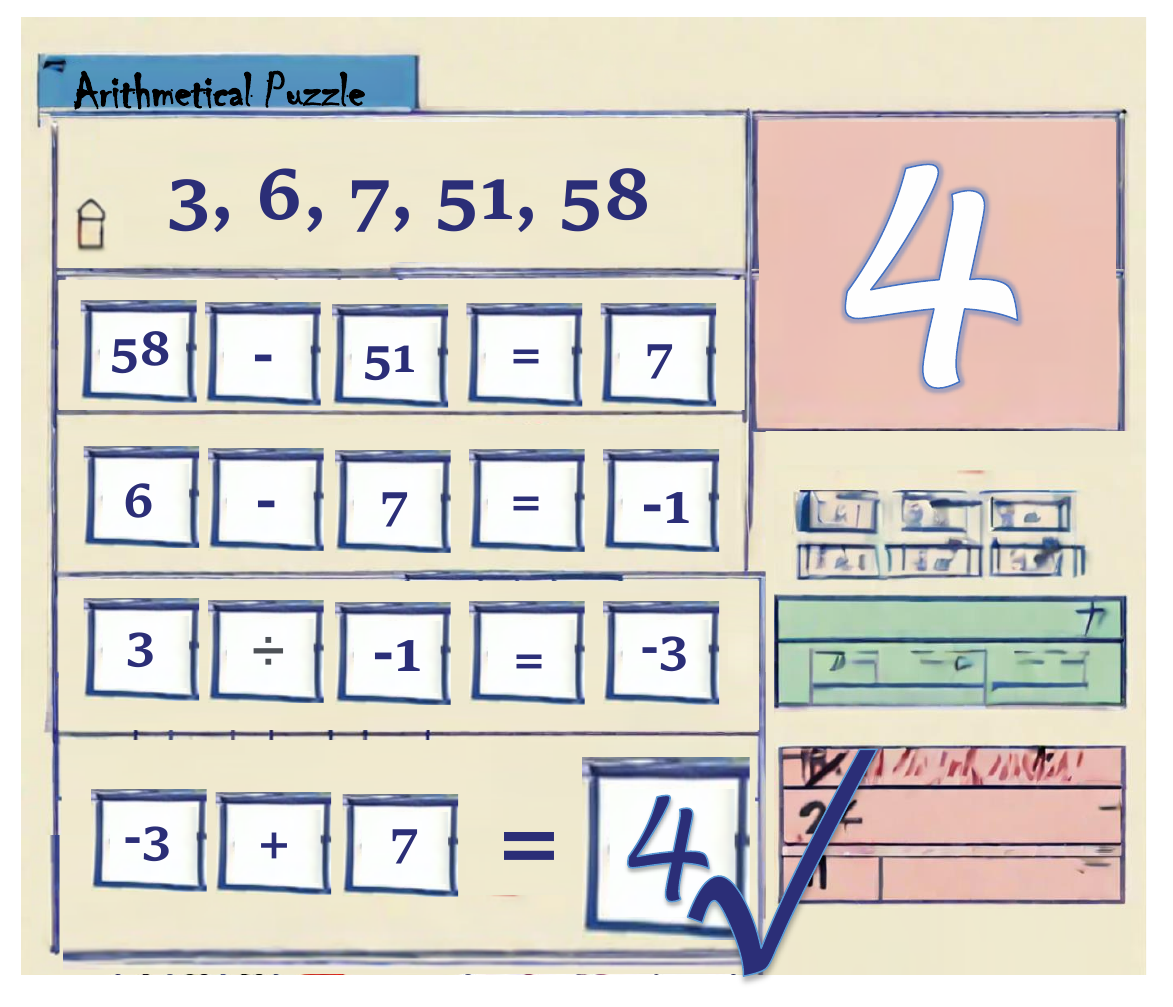}
    \caption{Visualization of the proposed arithmetical puzzle. Given the candidate integers $3, 6, 7, 51, 58$ and the target integer $4$, the answer is $58-51=7, 6-7=-1, 3\times(-1)=-3, -3+7=4$.}
    \label{fig:main}
\end{figure}

\end{document}